\documentclass{article}
\usepackage{arxiv}
\usepackage[utf8]{inputenc}
\usepackage[T1]{fontenc}    
\usepackage{newtxtext} 
\usepackage{url}  
\usepackage{booktabs}   
\usepackage{amsmath,amssymb,amsfonts}  
\usepackage{nicefrac}
\usepackage{microtype}    
\usepackage{multirow} 
\usepackage{lipsum}
\usepackage{graphicx}
\usepackage{natbib}
\usepackage{doi}
\usepackage{algorithm}
\usepackage{algpseudocode}
\usepackage{hyperref}

\title{A Hierarchical Computer Vision Pipeline for Physiological Data Extraction from Bedside Monitors}

\author{
  Vinh Chau\textsuperscript{1,*},  
  Khoa Le Dinh Van\textsuperscript{1},
  Hon Huynh Ngoc \textsuperscript{2},
  Binh Nguyen Thien\textsuperscript{2},
  Hao Nguyen Thien\textsuperscript{2},\\
  \bfseries
  Vy Nguyen Quang\textsuperscript{1},
  Phuc Vo Hong\textsuperscript{1}, 
  Yen Lam Minh \textsuperscript{1},
  Kieu Pham Tieu\textsuperscript{1},
  Trinh Nguyen Thi Diem\textsuperscript{1},\\
  \bfseries
  Louise Thwaites\textsuperscript{1,3}, 
  Hai Ho Bich\textsuperscript{1,3}\\
    \textsuperscript{1}Oxford University Clinical Research Unit, Ho Chi Minh City, Viet Nam \\
    \textsuperscript{2} Trung Vuong Hospital, Ho Chi Minh City, Vietnam \\
    \textsuperscript{3} Nuffield Department of Medicine, University of Oxford, United Kingdom \\
  \textsuperscript{*} Corresponding authors: \texttt{vinhc@oucru.org} \\
}

\begin{document}
\maketitle

\begin{abstract}

In many low-resource healthcare settings, bedside monitors remain standalone legacy devices without network connectivity, creating a persistent interoperability gap that prevents seamless integration of physiological data into electronic health record (EHR) systems. To address this challenge without requiring costly hardware replacement, we present a computer vision-based pipeline for the automated capture and digitisation of vital sign data directly from bedside monitor screens. Our method employs a hierarchical detection framework combining YOLOv11 for accurate monitor and region of interest (ROI) localisation with PaddleOCR for robust text extraction. To enhance reliability across variable camera angles and lighting conditions, a geometric rectification module standardizes the screen perspective before character recognition. We evaluated the system on a dataset of 6,498 images collected from open-source corpora and real-world intensive care units in Vietnam. The model achieved a mean Average Precision (mAP@50-95) of 99.5\% for monitor detection and 91.5\% for vital sign ROI localisation. The end-to-end extraction accuracy exceeded 98.9\% for core physiological parameters, including heart rate, oxygen saturation (SpO$_2$), and arterial blood pressure. These results demonstrate that a lightweight, camera-based approach can reliably transform unstructured information from screen captures into structured digital data, providing a practical and scalable pathway to improve information accessibility and clinical documentation in low-resource settings.

\end{abstract}


\section{Introduction}
\label{sec:intro}

In the critical care landscape of Low- and Middle-Income Countries (LMICs), continuous surveillance by bedside monitors is the cornerstone of patient safety. These devices generate high-frequency streams of physiological data—ranging from heart rate and blood pressure to oxygen saturation—that are vital for immediate clinical decision-making and longitudinal analysis of patient trajectories ~\cite{AbouZahr2005, Celi2013}. However, despite the ubiquity of these monitors, a critical ``interoperability gap'' persists, effectively trapping valuable data within the device's display ~\cite{JosephEjiyi2023}.

Currently, the bridge between bedside monitors and Electronic Health Records (EHRs) is predominantly manual transcription. This analog workflow is not only labor-intensive but also introduces significant latency and potential for error. Recent prospective studies have rigorously quantified these inefficiencies. Nitayavardhana et al. ~\cite{Nitayavardhana2025} demonstrated that transitioning from manual entry to an optical character recognition (OCR) workflow could reduce data recording time by approximately 44\%, significantly alleviating the administrative burden on clinical staff. Furthermore, Soeno et al. ~\cite{Soeno2024} highlighted the reliability of automated systems, reporting a near-zero error rate for OCR-based vital sign capture compared to the inevitable inaccuracies inherent in manual typing.

The challenge of digitization is particularly pronounced in resource-constrained settings, where healthcare facilities often rely on a diverse mix of legacy or donated equipment. As noted by ~\cite{Ferreira2025, Hinrichs-Krapels2022}, many of these medical devices operate as "closed systems," using proprietary communication protocols and lacking standardized interfaces for data export. Implementing centralized integration solutions in such environments is frequently prohibitively expensive or technically unfeasible. As a result, physiological data is often transient—displayed briefly before being discarded—leading to a significant loss of clinical intelligence.

Computer vision and deep learning have emerged as transformative technologies capable of bridging this gap non-invasively ~\cite{Jiang2017, Litjens2017}. By treating the monitor display as a visual signal source, camera-based approaches can digitize vital signs without requiring hardware modifications or vendor-specific drivers. Such systems hold transformative potential for LMICs, promising to enable remote multi-patient monitoring and real-time analytics using ubiquitous, low-cost imaging hardware like smartphones ~\cite{Topol2019}.

However, deploying these solutions in real-world ICUs presents unique computational and environmental challenges. Existing approaches often rely on cloud-based processing or computationally intensive models, which are ill-suited for edge deployment in settings with unstable internet connectivity. To address this, our research presents a robust, lightweight computational pipeline for the automated extraction of vital signs. Our methodology is specifically engineered to handle the complexities of the clinical environment—such as screen glare, oblique viewing angles, and diverse monitor layouts—while maintaining high inference efficiency suitable for deployment on modest hardware typical of LMIC settings.

\raggedbottom
\section{Related Work}

Recent reviews highlight that computer vision is widely used in healthcare for non-contact tasks, such as monitoring patient mobility and safety. However, the critical research gap in directly extracting physiological data from existing devices in real time remains underexplored, especially in resource-limited settings, emphasizing the urgent need for innovative solutions.

The majority of literature employs computer vision for indirect estimation, such as studies by Siegel et al. ~\cite{Siegel2024}, Yeung et al. ~\cite{Yeung2019}, and Davoudi et al. ~\cite{Davoudi2019}. These use depth and multi-modal sensors to assess mobility but depend on visual approximations, which lack the precision of bedside monitors, underscoring the need for more direct, accurate solutions.

Recent efforts in direct data extraction, such as those by Nitayavardhana et al.~\cite{Nitayavardhana2025}, have demonstrated that cloud-based OCR systems are validated across multiple ICUs. However, reliance on cloud infrastructure introduces latency and connectivity challenges, especially in LMIC environments, emphasizing the need for local, resource-efficient solutions. 

More recently, Chikhale and Mehendale ~\cite{Chikhale2025} introduced a framework linking vital sign digitization with an adaptive drug infusion system. Their approach uses Tesseract OCR, coupled with a fuzzy-logic controller, to automate medication delivery based on real-time readings. While they reported a high accuracy of 99.87\% on synthetic datasets, their reliance on traditional image processing techniques and simulated environments leaves the system's robustness unproven against the complex lighting artifacts and diverse screen layouts typical of real-world clinical settings. This highlights the opportunity for further innovation to improve robustness in real-world conditions.

Ferreira et al. ~\cite{Ferreira2025} conducted a comparative study of single-stage detectors for medical displays, utilizing a dataset sourced from Finnegan et al. ~\cite{Finnegan2019}. Crucially, this dataset consisted primarily of home-use devices (e.g., glucose meters), which typically feature simple, static seven-segment displays. These devices lack the visual complexity of the multi-parameter bedside monitors found in ICUs. Consequently, the applicability of such models to the dynamic, waveform-rich interfaces of ICU equipment remains limited.

Another important contribution is the work by Rampuria et al.~\cite{Rampuria2025}, which introduced a pipeline that uses EfficientNet-B4 for screen segmentation and PaddleOCR for information extraction. However, reliance on computationally intensive frameworks requires substantial hardware resources, creating a barrier to adoption in environments with limited infrastructure.

To address this gap, our study proposes a novel, lightweight State-of-the-Art (SOTA) pipeline tailored for ICU bedside monitors. Our method is designed to handle complex visual layouts efficiently, offering high accuracy and low latency on modest hardware, making it a promising solution for resource-limited settings.

\section{Materials and Methods}
\label{sec:methods}

\subsection{Overview of the Proposed Pipeline}

We formulate the vital sign extraction task as a hierarchical computer vision problem. Given an input image $I \in \mathbb{R}^{H \times W \times 3}$, the system outputs a structured set of measurements $D_{\text{out}} = \{(l_i, v_i)\}_{i=1}^{N}$, where $l_i$ is the semantic label (e.g., HR, SpO2) and $v_i$ is the numerical value. The pipeline is decomposed into three sequential stages (Algorithm ~\ref{alg:vital-extraction}):

\begin{itemize}
    \item \textbf{Monitor Localization}: We utilize a lightweight segmentation model to isolate the patient monitor screen from the background. This is followed by a perspective transformation to map the screen to a canonical view.
    \item \textbf{Vital Sign ROI Detection}: We employ an object detection model to identify specific Regions of Interest (ROIs) for each vital sign. This layout-agnostic approach allows the system to locate data regardless of screen position.
    \item \textbf{OCR and Data Digitization}: We extract raw text using an Optical Character Recognition (OCR) network. The raw output is then passed through a domain-specific logic validation filter to mitigate common OCR misinterpretations. This module applies two types of heuristics: syntactic correction, which uses Regex to rectify character confusion (e.g., identifying 'S' as '5' or 'O' as '0' in numerical fields), and physiological range gating, which discards values that are biologically impossible (e.g., SpO$_2 > 100$, Heart Rate $> 300$ or $< 10$). This post-processing step ensures that transient detection errors do not propagate into the final structured dataset.
\end{itemize}

\begin{algorithm}
\caption{Vital Sign Extraction Pipeline}
\label{alg:vital-extraction}
\begin{algorithmic}[1]
\Require Input image $I \in \mathbb{R}^{H \times W \times 3}$
\Require Models: $M_{\text{seg}}$ (Screen Seg), $M_{\text{det}}$ (ROI Det), $M_{\text{ocr}}$ (Text Rec)
\Require Confidence threshold $\tau = 0.8$
\Ensure Dictionary of vital signs $\mathcal{V} = \{\text{label}: [\{\text{val}, \text{conf}\}]\}$
\Statex
\Statex \textbf{\textit{Stage 1: Screen Localization \& Rectification}}
\State $\text{mask} \gets M_{\text{seg}}(I, \tau)$ 
\If{$\text{mask} = \varnothing$}
    \State \Return $\varnothing$ \Comment{Termination: Monitor not found}
\EndIf
\State $\mathcal{Q} \gets \textsc{ExtractCorners}(\text{mask})$ \Comment{Derive 4 corner points}
\State $I_{\text{rect}} \gets \textsc{PerspectiveWarp}(I, \mathcal{Q})$ \Comment{Align screen to standard view}
\Statex
\Statex \textbf{\textit{Stage 2: Region of Interest (ROI) Detection}}
\State $\text{boxes} \gets M_{\text{det}}(I_{\text{rect}}, \tau)$ \Comment{Detect classes (e.g., hr, spo2)}
\State $\mathcal{V} \gets \{\}$ \Comment{Initialize output dictionary}
\Statex
\Statex \textbf{\textit{Stage 3: OCR \& Data Digitization}}
\For{each box $b$ in $\text{boxes}$}
    \State $\text{label} \gets b.\text{class\_name}$
    \State $I_{\text{crop}} \gets \textsc{Crop}(I_{\text{rect}}, b.\text{bbox})$ \Comment{Isolate specific vital sign}
    \State $\text{ocr\_result} \gets M_{\text{ocr}}(I_{\text{crop}})$
    \If{$\text{ocr\_result} = \varnothing$}
        \State \textbf{continue}
    \EndIf
    \State $\text{text}_{\text{raw}} \gets \text{ocr\_result}.\text{text}$
    \State $\text{score} \gets \text{ocr\_result}.\text{score}$
    \State $\text{text}_{\text{clean}} \gets \textsc{Validate}(\text{text}_{\text{raw}}, \text{score})$ \Comment{Regex filter \& logic check}
    \If{$\text{text}_{\text{clean}} \neq \text{NULL}$}
        \State $\mathcal{V}[\text{label}].\text{append}(\{\text{val}: \text{text}_{\text{clean}}, \text{conf}: \text{score}\})$
    \EndIf
\EndFor
\State \Return $\mathcal{V}$
\end{algorithmic}
\end{algorithm}

\subsection{Data Collection and Preparation}
\label{sec:data_collection}

\subsubsection{Public Dataset with Annotations}
\label{subsec:public_data}
To ensure thorough coverage of various visual patterns, we incorporated publicly available data from the Roboflow Universe platform, specifically the Cloudphysician part 1 dataset ~\cite{Midprep2024}. The dataset is structured into two main components:
\begin{itemize}
    \item \textbf{Screen Detection Dataset:} Contains approximately \textbf{2,000} annotated images with a single class label: \textbf{`Screen'}.
    \item \textbf{Vital Sign Dataset:} Contains approximately \textbf{2,400} annotated images encompassing \textbf{8} distinct class labels, as detailed in Table ~\ref{tab:vital_sign_distribution}.
\end{itemize}

\begin{table}
\centering
\caption{Distribution of annotated instances in the Vital Sign Dataset, organized by physiological category.}
\label{tab:vital_sign_distribution}
\begin{tabular}{lr} 
\toprule
\textbf{Class Name} & \textbf{Instances} \\
\midrule
\multicolumn{2}{l}{\textit{Cardiac \& Oxygenation}} \\
HR (Heart Rate)              & 2,331 \\
PR (Pulse Rate)              & 1,220 \\
SpO$_2$ (Oxygen Saturation)  & 2,317 \\
\midrule
\multicolumn{2}{l}{\textit{Blood Pressure}} \\
SYS (Systolic BP)            & 2,211 \\
DIA (Diastolic BP)           & 2,260 \\
MAP (Mean Arterial Pressure) & 2,260 \\
\midrule
\multicolumn{2}{l}{\textit{Other Parameters}} \\
RR (Respiratory Rate)        & 2,256 \\
TEMP (Temperature)           & 940   \\
\midrule
\textbf{Total}               & \textbf{15,795} \\
\bottomrule
\end{tabular}
\end{table}

\subsubsection{Clinical In-house Dataset}
\label{subsec:oucru_data}
To represent the target domain, we collected 2,098 images from bedside monitors of various types, including GE, Philips IntelliVue, and Nihon Kohden, from the Intensive Care and Cardiology departments at Trung Vuong Hospital in Vietnam. The study received ethical approval from the hospital ethics committee to recruit 100 patients who may require escalated care. The study team captured bedside monitor screens daily with mobile phones in authentic settings; this dataset introduces critical real-world complexities, including extreme lighting variations (glare), partial occlusions, and device heterogeneity. To ensure data quality, annotation was performed by clinicians and subsequently cross-verified.

\subsection{Monitor Localization and Vital Sign Detection}

\subsubsection{Model Architecture}
We employ YOLOv11 ~\cite{Khanam2024, Jegham2025}, a state-of-the-art, one-stage object detection architecture. Released in late 2024, YOLOv11 introduces C3k2 Bottleneck Blocks for parameter efficiency and C2PSA (Cross-Stage Partial with Spatial Attention) to enhance detection robustness under challenging visual conditions ~\cite{Khanam2024, Jegham2025}.

\subsubsection{Model Selection Strategy}
Our pipeline adopts a heterogeneous model strategy:
\begin{itemize}
    \item \textbf{Stage 1 (YOLOv11n-seg):} For monitor localization, we prioritized the nano variant (2.8M parameters) to ensure real-time processing on CPU-only devices.
    \item \textbf{Stage 2 (YOLOv11s):} For vital sign extraction, we employed the small variant (9.4M parameters). The increased capacity is necessary to resolve fine-grained details between visually similar values (e.g., SYS vs. DIA).
\end{itemize}

\subsection{Geometric Rectification}
\label{subsec:geometric_rectification}
To ensure robustness against varying camera angles, we implement a geometric normalization module. Using the segmentation mask $m^*$, we approximate the monitor boundary using the Douglas-Peucker algorithm or a minimum-area rectangle. The source points $P_{\mathrm{src}}$ are mapped to a standardized destination coordinate system $P_{\mathrm{dst}}$ ($640 \times 480$ pixels) via a homography matrix $H$:
\begin{equation}
    \begin{bmatrix} x'_i \\ y'_i \\ 1 \end{bmatrix} \sim H \begin{bmatrix} x_i \\ y_i \\ 1 \end{bmatrix}
\end{equation}
The rectified image $I_{\mathrm{rect}}$ is generated by applying an inverse perspective warp.

\subsection{OCR-based Data Extraction via PP-OCRv5}
\label{subsec:ocr_extraction}
For the core recognition task, we employ the \textbf{PP-OCRv5} Mobile model from the PaddleOCR framework ~\cite{Cui2025}. This architecture was selected for its balance between inference speed and accuracy. It features a text detection module based on PP-HGNetV2, a direction classifier for orientation correction, and a text recognition head using SVTR-HGNet. Therefore, to achieve real-time performance on edge devices, we utilized the mobile version of the architecture.

\section{Experiments and Results}
\label{sec:results}

\subsection{Experimental Setup}
\label{subsec:setup}
The annotated public dataset was partitioned into \textbf{70\% for training, 20\% for validation, and 10\% for testing}. The real-world clinical dataset ($N=2,098$) was reserved exclusively for external testing. Models were implemented in PyTorch and trained on an NVIDIA A100 (40GB VRAM). Inference benchmarking was conducted on three hardware tiers:
\begin{itemize}
    \item \textbf{Config A (Baseline):} Intel Core i7-11700 (CPU Only).
    \item \textbf{Config B (Edge/Mobile):} Intel Core i5-1135G7 (Iris Xe Graphics).
    \item \textbf{Config C (Workstation):} Intel Core i7-10700 + NVIDIA GTX 1660 Ti.
\end{itemize}

\subsection{Monitor Localization \& Rectification}
The YOLOv11n-seg model achieved robust convergence (Supplementary Fig.~\ref {fig:supp_training_curves}). On the validation set (Table~\ref {tab:test_metrics}), it attained a Mask mAP@50-95 of 98.6\%. On the pixel-level test set, the model achieved a Dice coefficient of $0.9867 \pm 0.0048$ (Table ~\ref{tab:test_pixel_metrics}).

\begin{table}
    \centering
    \caption{Validation performance of the YOLOv11n-seg model.}
    \label{tab:test_metrics}
    \setlength{\tabcolsep}{12pt}
    \begin{tabular}{lcccc}
        \toprule
        \textbf{Task} & \textbf{Precision} & \textbf{Recall} & \textbf{mAP@50} & \textbf{mAP@50-95} \\ 
        \midrule
        Bounding Box & 0.998 & 0.999 & 0.995 & 0.995 \\
        Segmentation Mask & 0.999 & 0.997 & 0.995 & 0.986 \\ 
        \bottomrule
    \end{tabular}
\end{table}

\begin{table}
\centering
\caption{Pixel-level segmentation metrics on the test set ($N=200$).}
\label{tab:test_pixel_metrics}
\setlength{\tabcolsep}{12pt}
\begin{tabular}{lc}
\toprule
\textbf{Metric} & \textbf{Score (Mean $\pm$ Std)} \\
\midrule
IoU         & $0.9738 \pm 0.0093$ \\
Dice        & $0.9867 \pm 0.0048$ \\
Precision   & $0.9801 \pm 0.0098$ \\
Recall      & $0.9938 \pm 0.0047$ \\
\bottomrule
\end{tabular}
\end{table}

In real-world benchmarking (Table~\ref {tab:inference_benchmark}), Config C demonstrated real-time capability at 81.3 FPS with a 99.8\% detection success rate.

\begin{table}
\centering
\caption{Inference benchmark on the OUCRU clinical dataset ($N=2{,}098$).}
\label{tab:inference_benchmark}
\begin{tabular}{lcccc}
\toprule
\multirow{2}{*}{\textbf{Configuration}} & \multicolumn{2}{c}{\textbf{Efficiency}} & \multicolumn{2}{c}{\textbf{Reliability}} \\ 
\cmidrule(lr){2-3} \cmidrule(lr){4-5}
 & \textbf{Latency (ms)} & \textbf{FPS} & \textbf{Avg. Conf.} & \textbf{Success (\%)} \\ 
\midrule
Config A (CPU) & 75.7 & 13.2 & 0.926 & 98.4 \\
Config B (Mid-range) & 49.5 & 20.2 & 0.919 & 98.3 \\
Config C (GPU) & 12.3 & 81.3 & 0.921 & 99.8 \\ 
\bottomrule
\end{tabular}
\end{table}

\subsection{Vital Sign Region of Interest (ROI) Detection}
The quantitative evaluation on the test set demonstrates the robustness of the YOLOv11s architecture. As detailed in Table \ref{tab:roi_test_perclass}, the model achieved an aggregate mAP@50 of 0.993 with consistent performance across diverse physiological parameters. Notably, the high mAP@50-95 scores (mean 0.915) indicate precise bounding-box regression. These metrics confirm the model's generalization capability on independent test data, ensuring that vital signs are accurately localized regardless of minor variations in display layouts.

\begin{table}
\centering
\caption{Per-class detection performance on the test set. Consistent high performance is observed across all physiological groups.}
\label{tab:roi_test_perclass}
\setlength{\tabcolsep}{8pt}
\begin{tabular}{lcccc}
\toprule
\textbf{Vital Sign} & \textbf{Precision} & \textbf{Recall} & \textbf{mAP@50} & \textbf{mAP@50-95} \\ 
\midrule
HR (Heart Rate)              & 0.983 & 0.996 & 0.995 & 0.953 \\
PR (Pulse Rate)              & 0.978 & 0.974 & 0.993 & 0.893 \\
SpO$_2$ (Oxygen Saturation)  & 0.983 & 0.987 & 0.991 & 0.955 \\
SYS (Systolic BP)            & 0.982 & 0.987 & 0.993 & 0.944 \\
DIA (Diastolic BP)           & 0.999 & 1.000 & 0.995 & 0.936 \\
MAP (Mean Arterial Pressure) & 0.989 & 1.000 & 0.995 & 0.936 \\
RR (Respiratory Rate)        & 0.978 & 0.974 & 0.987 & 0.928 \\
TEMP (Temperature)           & 0.989 & 0.993 & 0.994 & 0.776 \\
\midrule
\textbf{Mean}                & \textbf{0.985} & \textbf{0.989} & \textbf{0.993} & \textbf{0.915} \\
\bottomrule
\end{tabular}
\end{table}

\begin{figure}[htbp]
    \centering
    \includegraphics[width=0.7\linewidth]{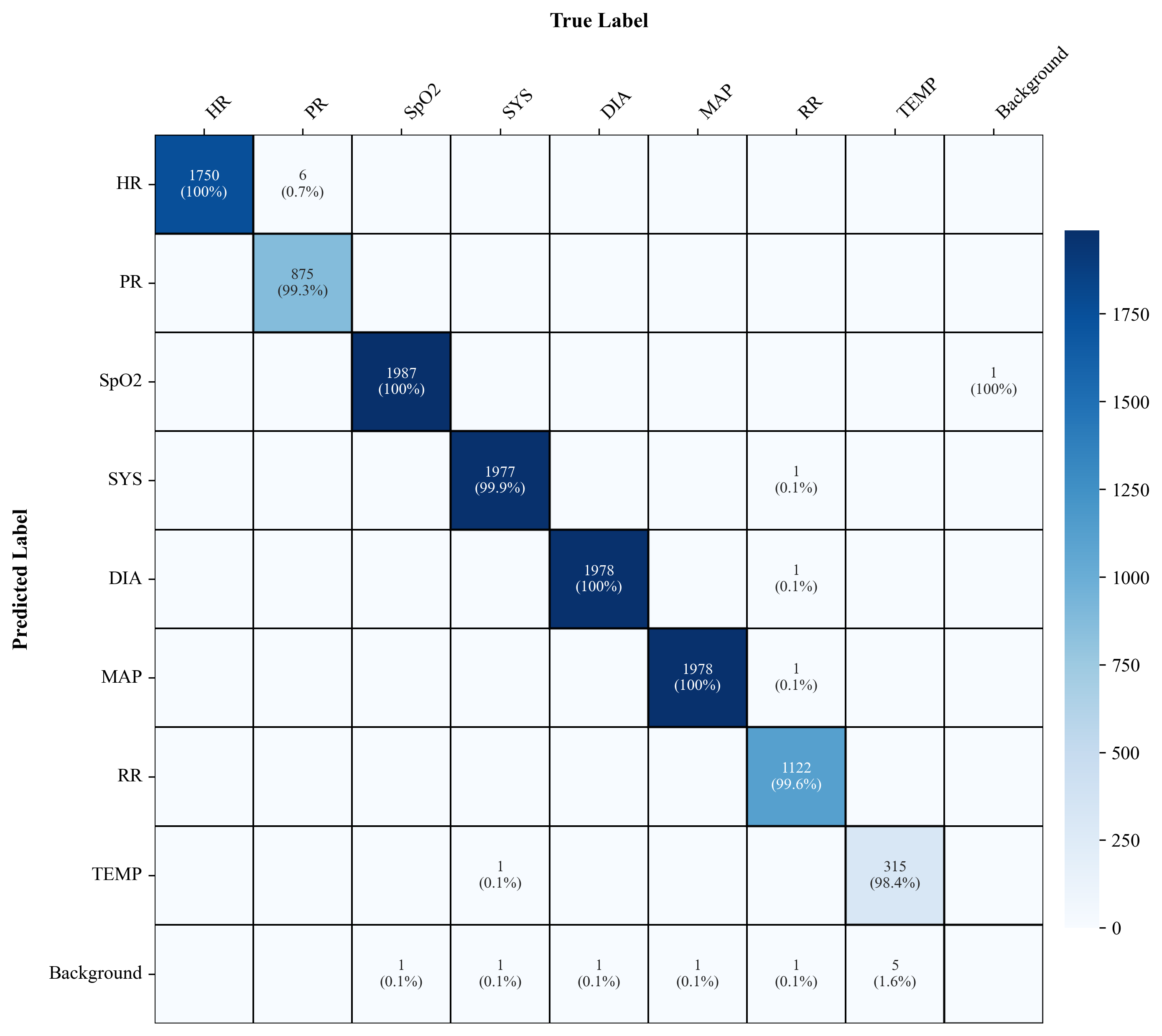}
    \caption{Confusion matrix of the ROI detection model evaluated on the OUCRU Clinical Dataset ($N=12,002$ instances).}
    \label{fig:confusion_matrix}
\end{figure}

To rigorously analyze the classification behavior and error patterns in a real-world clinical setting, we computed the Confusion Matrix over the entire OUCRU Clinical Dataset comprising 12,002 ROI instances (Figure~\ref{fig:confusion_matrix}). In this visualization, rows correspond to the predicted classes, and columns represent the ground-truth labels. The matrix exhibits a pronounced diagonal dominance, corroborating the model's high precision across major physiological categories. However, a granular examination reveals two specific, albeit minor, error modes. First, semantic ambiguity was observed between Heart Rate (HR) and Pulse Rate (PR). Specifically, 0.68\% of ground-truth PR instances were misclassified as HR. This is attributable to the visual similarity in font properties and color schemes often employed for these parameters on legacy monitoring displays. Second, regarding false negatives, the Temperature (TEMP) class proved the most susceptible to occlusion. The model failed to localize 1.56\% of TEMP instances, likely due to the smaller spatial footprint of temperature indicators. Despite these edge cases, the system demonstrated exceptional specificity, with a negligible overall missed-detection rate (approximately 0.08\% across 12,002 instances) and virtually zero false positives in background regions.

Beyond detection accuracy, operational feasibility was assessed by measuring inference latency on the OUCRU clinical dataset to verify suitability for continuous monitoring. On the workstation configuration (Config C), the ROI detection module averaged 16.9 ms per frame. Conversely, in the resource-constrained CPU environment (Config A), latency increased to 79.1 ms per frame. These metrics indicate that the YOLOv11s model effectively balances architectural complexity with computational efficiency, satisfying the real-time constraints required for continuous multi-vital sign monitoring without imposing a prohibitive bottleneck.

\subsection{OCR-based Data Extraction}
The OCR module was evaluated on the OUCRU clinical dataset. Table~\ref {tab:ocr_accuracy} details the extraction accuracy. The system achieved an overall accuracy of 99.08\% across 12,002 total instances. SpO$_2$ showed the highest reliability (99.62\%), while HR was slightly lower (98.9\%), likely due to the dynamic updating frequency of heart rate values causing minor motion blur.

\begin{table}
\centering
\caption{OCR extraction accuracy on the OUCRU clinical dataset ($N=2{,}098$ images per class). Vital signs are grouped by clinical relevance.}
\label{tab:ocr_accuracy}
\setlength{\tabcolsep}{6pt}
\begin{tabular}{lccc}
\toprule
\textbf{Vital Sign} & \textbf{Correct Matches} & \textbf{Total Instances} & \textbf{Field-level accuracy (\%)} \\ 
\midrule
HR (Heart Rate)              & 1,730 & 1,750 & 98.9 \\
PR (Pulse Rate)              & 864   & 881   & 98.1 \\
SpO$_2$ (Oxygen Saturation)  & 1,980 & 1,988 & 99.6 \\
SYS (Systolic BP)            & 1,963 & 1,979 & 99.2 \\
DIA (Diastolic BP)           & 1,961 & 1,979 & 99.1 \\
MAP (Mean Arterial Pressure) & 1,969 & 1,979 & 99.5 \\
RR (Respiratory Rate)        & 1,111 & 1,126 & 98.7 \\
TEMP (Temperature)           & 314   & 320   & 98.2 \\
\midrule
\textbf{Overall}             & \textbf{11,892} & \textbf{12,002} & \textbf{99.08} \\
\bottomrule
\end{tabular}
\end{table}

Regarding computational efficiency, the text recognition stage is the most intensive component of the pipeline because it requires processing multiple cropped regions sequentially for each frame. Benchmarking on the OUCRU dataset yielded an average processing time of \textbf{37.6 ms} per image with the GPU-accelerated setup (Config C). On the standard CPU configuration (Config A), the inference time increased to \textbf{222.9 ms} per image. Although this stage incurs the highest latency, the aggregate speed remains sufficient to capture physiological changes, which typically occur at a frequency lower than the system's throughput.

\subsection{End-to-End Pipeline Performance}
\label{subsec:end_to_end_results}

We evaluated the comprehensive performance of the proposed pipeline by aggregating the success rates across all physiological categories. The system demonstrated a robust mean extraction accuracy of 98.9\% across the eight target vital signs. This high level of precision confirms that integrating YOLOv11 for detection and PaddleOCR for recognition creates a cohesive framework in which the geometric rectification step effectively bridges the gap between raw visual inputs and structured data outputs.

To characterize the overall latency of the proposed system, we aggregated the inference times across all three sequential stages—monitor localization, ROI detection, and OCR extraction—while accounting for a constant computational overhead of approximately 15 ms to cover image preprocessing, geometric rectification, and logic validation. On the high-performance workstation configuration (Config C), the cumulative processing latency per frame is approximately 81.8 ms (comprising 12.3 ms for localization, 16.9 ms for detection, 37.6 ms for OCR, and 15 ms overhead). This results in a throughput of $\sim$12.2 FPS, ensuring seamless real-time processing and immediate data digitization suitable for high-acuity monitoring scenarios.

Crucially, the evaluation demonstrates that the system remains operationally viable even on the baseline CPU-only hardware (Config A), which represents the most common infrastructure in resource-constrained settings. On this configuration, the total pipeline latency is approximately 392.7 ms (summing 75.7 ms, 79.1 ms, 222.9 ms, and the 15 ms overhead), yielding a throughput of roughly 2.5 FPS. In clinical practice, physiological parameters such as heart rate, blood pressure, and SpO$_2$ exhibit relatively slow temporal dynamics compared to video frame rates. Consequently, a sampling frequency of 2.5 Hz is sufficient to capture clinically significant updates. While this frame rate is lower than the standard video playback rate, it exceeds the Nyquist rate for vital sign trends, which typically evolve over seconds or minutes. This confirms that our solution enables effective automated charting and data logging on standard, low-cost hospital computers without requiring expensive GPU upgrades, thereby satisfying the core objective of accessibility for LMICs.

\section{Discussion}
\label{sec:discussion}

The primary contribution of this study is the development of a hierarchical, geometry-aware pipeline that effectively bridges the ``interoperability gap'' in resource-constrained ICUs. By integrating a lightweight YOLOv11 detector with a rectified OCR stage, our system demonstrates significant performance gains over existing methods. Although differences in test datasets constrain a direct quantitative comparison, the contrast in extraction fidelity offers valuable insights into architectural efficacy. While Rampuria et al. \cite{Rampuria2025}, utilizing an end-to-end approach (PaddleOCR + EfficientNetB2), reported accuracy peaks of 67.48\% for SpO$_2$ and 65.27\% for MAP, our system achieved an extraction accuracy exceeding 98.9\% across these same categories. We hypothesize that this substantial performance gap underscores the critical necessity of our geometric rectification module. Unlike baseline approaches that attempt to recognize text directly from raw images, often with skewed camera angles, our method neutralizes perspective distortions—a primary source of OCR failure in real-world bedside monitoring—before the recognition stage. This canonical alignment effectively standardizes the visual input, ensuring reliable digitization even under oblique viewing conditions.

Beyond quantitative accuracy, this approach offers critical practical advantages for healthcare infrastructure in Low- and Middle-Income Countries (LMICs). As highlighted by Ferreira et al. ~\cite{Ferreira2025}, replacing legacy, non-networked bedside monitors with modern connected devices is often financially unfeasible. Our solution circumvents this barrier by treating the monitor display as a universal interface, enabling the digitization of ``closed'' systems regardless of the manufacturer (e.g., Philips, GE, Nihon Kohden). Furthermore, by leveraging computer vision via existing CCTV or commodity smartphones, the system facilitates non-contact monitoring. This reduces the need for direct physical interaction with the device, thereby minimizing workflow disruptions and enhancing infection control protocols in the ICU.

We acknowledge several limitations in our current study design. First, while our dataset is derived from authentic clinical environments, its total volume is modest compared to that of large-scale industrial datasets. A notable class imbalance persists, particularly regarding Temperature (TEMP) readings. Furthermore, visual analysis reveals that TEMP values inherently present a harder detection challenge: they are typically displayed in smaller fonts compared to HR/SpO$_2$, are often located in peripheral screen corners, and are frequently occluded by cabling or stickers. These physical factors, combined with limited training samples (940 instances), contributed to the lower localization consistency (mAP@50-95 of 77.6\%) for this class. Second, to prioritize real-time inference on edge devices, we utilized the mobile version of the PP-OCRv5 model. While efficient, this general-purpose engine is not explicitly fine-tuned for the specific dot-matrix or seven-segment LED fonts standard on older medical displays, potentially limiting character-level precision in extreme lighting conditions.

Finally, the current iteration of our pipeline is limited to extracting numerical data. We recognize that physiological waveforms (e.g., ECG, PPG) convey rich diagnostic information essential for detecting arrhythmias or perfusion anomalies. The digitization of these continuous signal streams remains a complex challenge not addressed in this work. Future development will focus on integrating a 1D signal reconstruction module to extract waveforms and expanding the dataset to include a broader diversity of monitor layouts, ultimately aiming to provide a comprehensive, fully digitized electronic medical record from visual sources.

\section{Conclusion}
\label{sec:conclusion}

This study presents a robust, end-to-end computer vision framework capable of digitizing physiological data from legacy ICU monitors with high fidelity. By integrating the lightweight YOLOv11 architecture for precise ROI localization with a geometry-aware OCR pipeline, we achieved an aggregate extraction accuracy exceeding 99\% across key vital signs, significantly outperforming existing baseline methods on challenging real-world data. This solution effectively addresses the critical interoperability gap in resource-constrained healthcare settings, offering a scalable, non-invasive alternative to costly hardware replacements.

Looking ahead, our future research will focus on transitioning this prototype into a deployable Edge-AI solution. Specifically, we aim to optimize the model quantization and pruning to facilitate direct deployment on embedded IoT devices such as the NVIDIA Jetson family (Nano/Orin) or Raspberry Pi ecosystems equipped with AI accelerators. This integration will enable the creation of standalone, low-power "smart camera" nodes that process data locally at the bedside, thereby ensuring privacy and reducing bandwidth requirements. Concurrently, we plan to develop a native mobile application to empower healthcare workers with point-of-care digitization capabilities via standard smartphones.

Furthermore, we intend to expand the pipeline's analytical depth by incorporating continuous waveform digitization (e.g., ECG and PPG signals), thereby providing a holistic view of patient health. Finally, to further enhance semantic understanding, we plan to investigate Transformer-based object detectors such as RT-DETRs (Real-Time Detection Transformer). Unlike CNN-based approaches, RT-DETR leverages global attention mechanisms better to capture the spatial context between labels and numerical values. This capability is expected to further minimize misclassification in highly cluttered screen layouts, paving the way for a fully autonomous, context-aware ICU monitoring system.

\bibliographystyle{unsrt}  
\bibliography{references}

\newpage
\section*{Supplementary Material}

\setcounter{figure}{0} 
\renewcommand{\thefigure}{S\arabic{figure}}

\begin{figure}[htbp]
    \centering
    \includegraphics[width=0.85\linewidth]{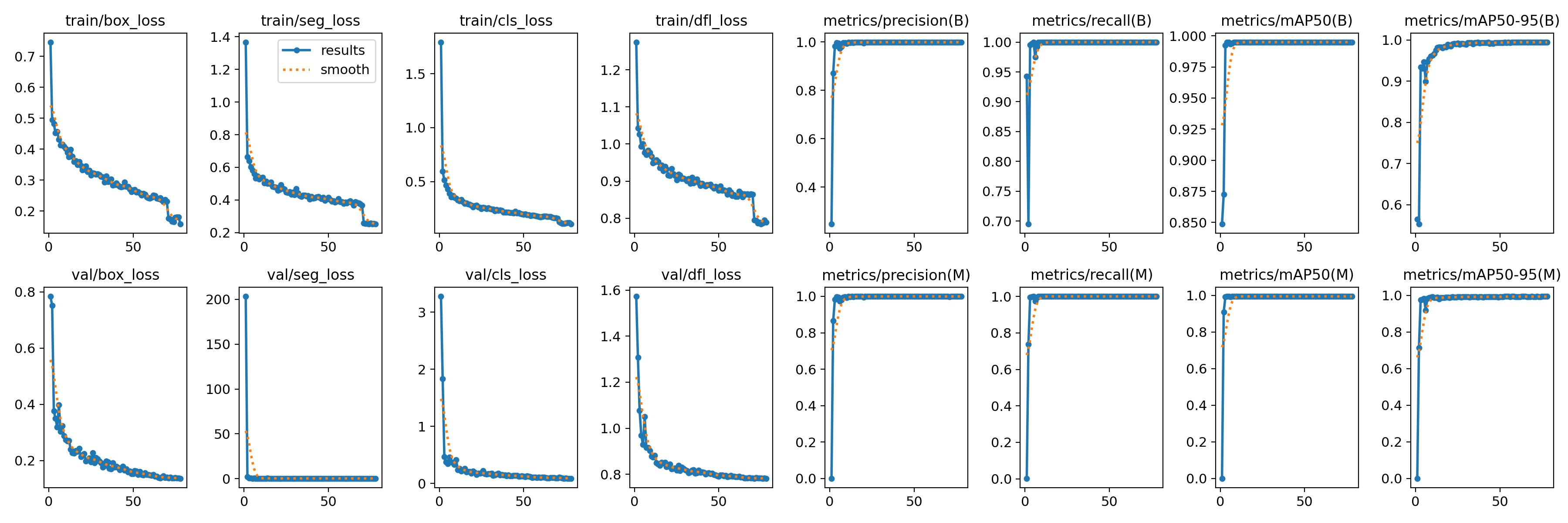} 
    \caption{Training and validation loss curves for the Monitor Localization model (YOLOv11n-seg)}
    \label{fig:supp_training_curves}
\end{figure}

\begin{figure}[htbp]
    \centering
    \includegraphics[width=0.85\linewidth]{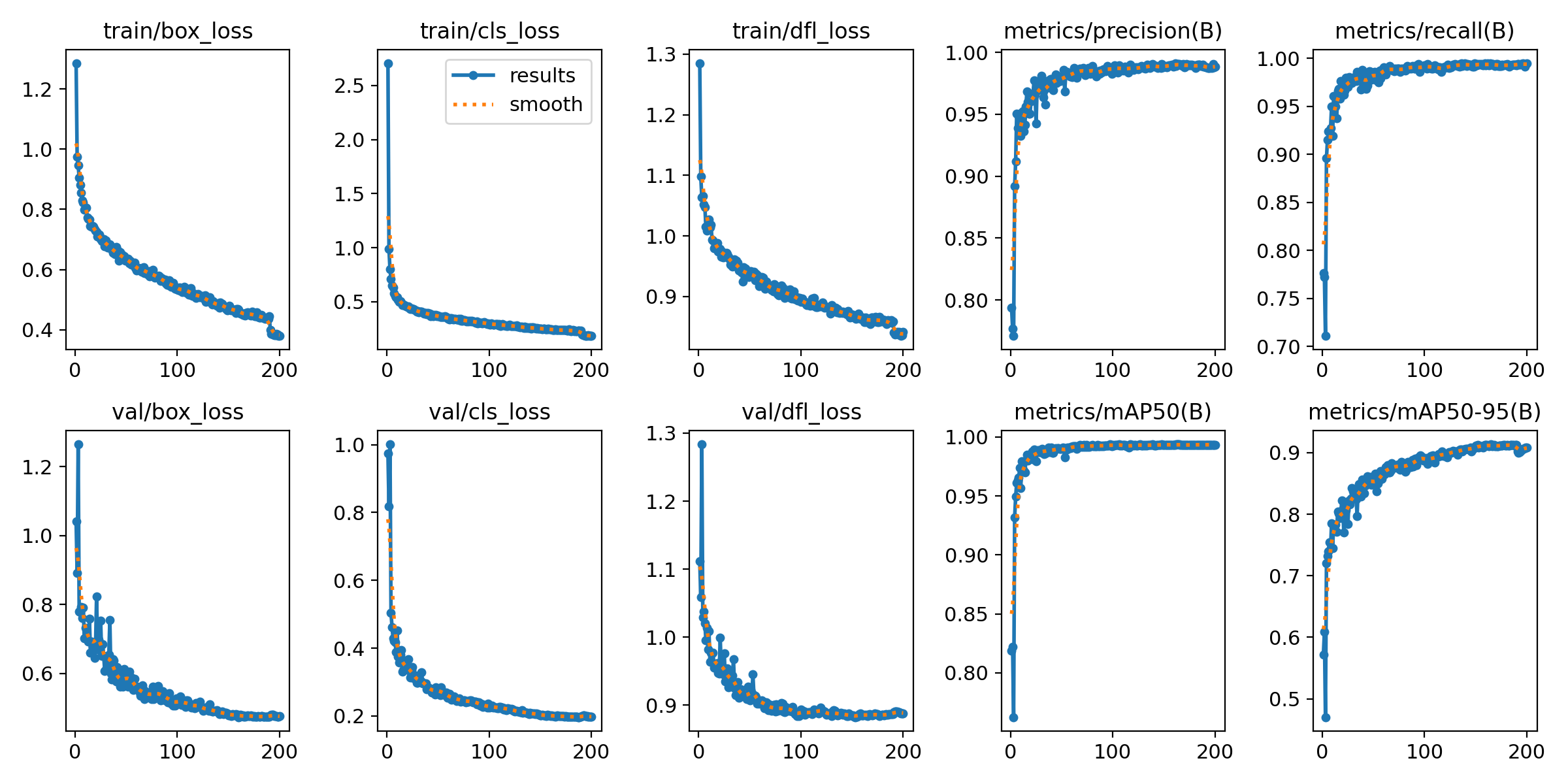} 
    \caption{Training dynamics of the YOLOv11s architecture for Vital Sign ROI detection.}
\label{fig:supp_roi_training_curves}
\end{figure}

\end{document}